\documentclass[letterpaper, 10 pt, conference]{IEEEconf}
\IEEEoverridecommandlockouts

\newcommand{\revision}[1]{{\color{black}#1}}
\usepackage{comment}
\usepackage{soul}
\usepackage{cite}
\usepackage{amsmath,amssymb,amsfonts}
\usepackage{algorithmic}
\usepackage{graphicx}
\usepackage{textcomp}
\usepackage{xcolor}
\usepackage{color}
\usepackage{multirow}
\usepackage{array}
\usepackage{threeparttable}
\usepackage{subcaption}
\usepackage{booktabs}
\usepackage[bookmarks=false]{hyperref}
\usepackage{cleveref}
\def\BibTeX{{\rm B\kern-.05em{\sc i\kern-.025em b}\kern-.08em
    T\kern-.1667em\lower.7ex\hbox{E}\kern-.125emX}}

\title{\LARGE \bf DRIFT: Dual-Representation Inter-Fusion Transformer for Automated Driving Perception with 4D Radar Point Clouds}

\author{Siqi Pei \\ \textit{Delft University of Technology} \\ Delft, The Netherlands
   \and Andras Palffy \\ \textit{Perciv AI} \\ Delft, The Netherlands
   \and Dariu M. Gavrila \\ \textit{Delft University of Technology} \\ Delft, The Netherlands
   }

\begin{document}
\maketitle
\pagestyle{empty}
\thispagestyle{empty}

\begin{abstract}
4D radars, which provide 3D point cloud data along with Doppler velocity, are attractive components of modern automated driving systems due to their low cost and robustness under adverse weather conditions. However, they provide a significantly lower point cloud density than LiDAR sensors. This makes it important to exploit not only local but also global contextual scene information.
%
%
This paper proposes DRIFT, a model that effectively captures and fuses both local and global contexts through a dual-path architecture. The model incorporates a point path to aggregate fine-grained local features and a pillar path to encode coarse-grained global features. These two parallel paths are intertwined via novel feature-sharing layers at multiple stages, enabling full utilization of both representations. DRIFT is evaluated on the widely used View-of-Delft (VoD) dataset \cite{apalffy2022} and a proprietary internal dataset. It outperforms the baselines on the tasks of object detection and/or free road estimation. For example, DRIFT achieves a mean average precision (mAP) of 52.6\% (compared to, say, 45.4\% of CenterPoint \cite{Yin2020} ) on the VoD dataset.
\end{abstract}
  
\section{Introduction}
\label{sec:introduction}

Automated driving systems have the potential to reduce traffic accidents, improve traffic efficiency, and provide mobility to people who cannot drive on their own. Cameras and LiDARs are widely studied sensors in automated driving, offering rich scene information. However, they face limitations: cameras struggle in low-light conditions and adverse weather like rain and fog, while LiDARs are expensive and can also be affected by these conditions \cite{Zhang_2023}, which hinders the performance of automated driving systems. In contrast, radar sensors are more robust under such conditions and also provide Doppler velocity and radar cross-section (RCS) information. Moreover, their significantly lower cost compared to LiDARs makes them an attractive alternative.

\begin{figure}[t]
  \centering
  \includegraphics[width=\linewidth, trim=0 60 0 50, clip]{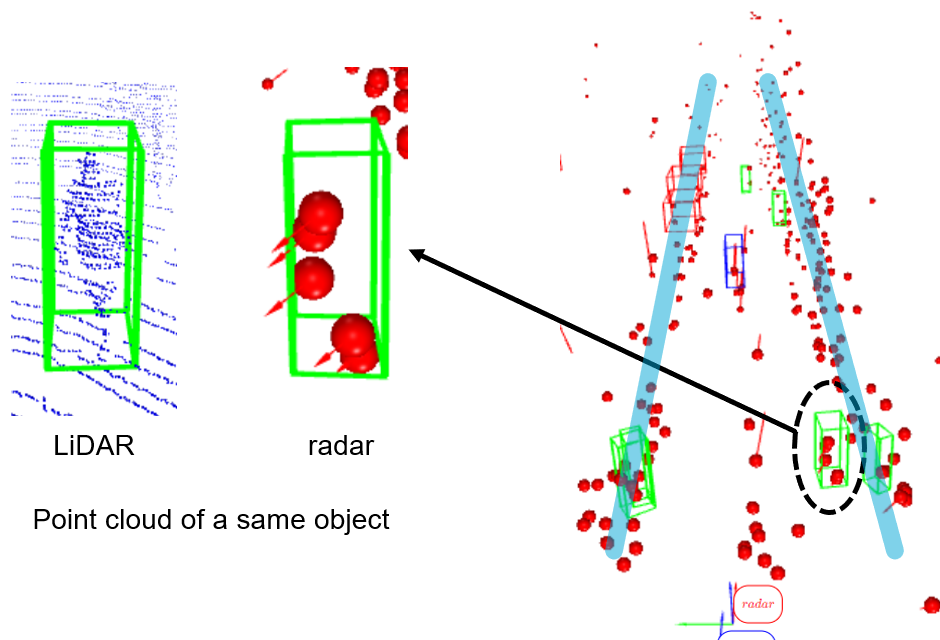}
  \caption{Example scene from the View-of-Delft \cite{apalffy2022} dataset with radar and annotation data. The two inlets on the left show corresponding LiDAR and radar point clouds around the same pedestrian. Unlike the LiDAR point cloud, it is difficult to detect the pedestrian using only the local region of the radar point cloud due to its sparsity. However, incorporating global information (i.e. the pedestrian's relative position to the ego-vehicle and the scene, e.g. drivable area marked with blue) with local features such as shape and velocity, the pedestrian's presence becomes more apparent.
  }

  \label{fig:example_long_range_info}
\end{figure} 

\begin{figure*}[t]
  \centering
  \begin{subfigure}{0.32\linewidth}
    \includegraphics[width=\linewidth]{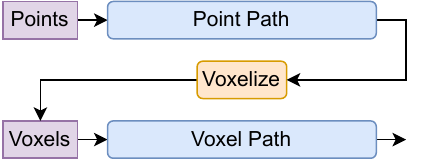}
    \caption{Sequential.}
    \label{fig:sequential}
  \end{subfigure}  
  \hfill
  \begin{subfigure}{0.32\linewidth}
    \includegraphics[width=\linewidth]{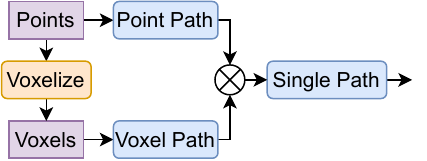}
    \caption{Parallel with fusion at the end.}
    \label{fig:parallel_fusion}
  \end{subfigure}
  \hfill
  \begin{subfigure}{0.32\linewidth}
    \includegraphics[width=\linewidth]{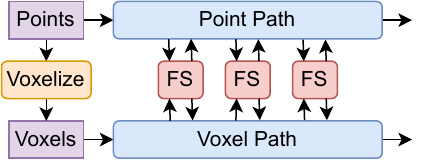}
    \caption{Parallel with feature sharing. (Proposed)}
    \label{fig:parallel_feature_sharing}
  \end{subfigure}
  \caption{Comparison of dual-representation model structures. FS denotes feature sharing block. (a) The sequential path processes the point cloud through a point-based path first, then through a voxel-based path, e.g. \cite{Leng2024}. (b) The parallel path with fusion at the end processes both paths independently and merges their outputs before proceeding with a single-representation path, e.g. \cite{Liu2019, JLiu2023}. (c) Proposed structure processes both paths in parallel and introduces feature sharing at each intermediate stage between them.}
  \label{fig:dual_representation_models}
\end{figure*}

Despite these advantages, radar sensors have certain drawbacks. Their lower resolution leads to much sparser point clouds compared to LiDARs. In addition, radar point clouds tend to contain more noise and clutter. As illustrated in \autoref{fig:example_long_range_info}, these drawbacks make understanding a scene from radar data require both local and global information, unlike LiDAR data, where local features are often sufficient. These challenges make it difficult for radar-based deep learning models to perform accurate perception, rendering LiDAR-based models less directly applicable to radar data.

For point cloud models, the most commonly used data representations are point-based and voxel/pillar-based. Point-based representations take the direct output of sensors as an $n\times(3+f)$ tensor, where $n$ is the number of points in the point cloud. Each point contains 3D coordinates $(x, y, z)$ and $f$ additional features \cite{Qi2017}. Voxel/Pillar-based representations divide the space into a 3D or 2D grid and aggregate the point features within each voxel into a single feature vector \cite{Zhou2017,Lang2018}. This process, known as voxelization, converts irregular point cloud data into regular grid data, allowing the application of convolutional methods typically used for images. However, it also causes loss of spatial detail and needs to tackle the sparsity problem \cite{Yan2018,Graham2017}. In radar models, pillar-based representations are preferred over voxel-based ones \cite{Musiat2024,apalffy2022} due to the sparsity of radar point clouds, which would otherwise result in a large number of empty voxels, making it difficult to extract meaningful features.

While traditional methods typically adopt either point-based or voxel-based representations, some recent works attempt to combine both in order to address better the sparse and noisy nature of 4D radar point clouds \cite{Peng2024,JLiu2023,Leng2024}. The point-based path operates directly on raw point-level data, enabling it to capture fine-grained local details. Meanwhile, the voxel-based path processes data at a coarser level, efficiently modeling global context. \autoref{fig:dual_representation_models} illustrates a comparison of different dual-representation model structures.

However, existing dual-representation models do not fully exploit the strengths of the two data representations. Point-based operations fail to utilize the high-level voxel features, and voxel-based operations overlook the fine-grained point-level details. Intermediate features within each path are often ignored, despite their potential usefulness for subsequent processing \cite{Lin2017}.

To address these limitations, a backbone architecture featuring end-to-end parallel \textit{point} and \textit{voxel} (pillar) paths is proposed in this paper, as shown in \autoref{fig:parallel_feature_sharing}. The model incorporates feature sharing blocks between intermediate stages in the two paths, enabling bi-directional feature exchange. This design allows both local and global information to be utilized at each stage, enhancing the effectiveness of feature extraction and fusion. Moreover, given the extreme sparsity of radar point clouds, relying solely on local or neighborhood information is insufficient to interpret scene content. Therefore, the model incorporates transformer-based layers into the voxel path, effectively extending the field of view and enabling the voxel path to model global dependencies even at early stages. This is feasible only for radar point clouds, as LiDAR point clouds are typically too dense for real-time global transformer applications. Transformers are also applied in the point path, enhancing its ability to capture relative local positional information. 
To fully exploit radars' sparsity, the entire model is implemented using sparse data representations. 
The contributions of this work are summarized as follows:
\begin{itemize}
    \item A novel dual-representation backbone architecture is proposed explicitly for radar point cloud processing. It features end-to-end parallel point and pillar paths, and is equipped with transformer-based layers and implemented entirely with sparse data representation.
    \item Novel feature sharing blocks are introduced at each intermediate stage, allowing bi-directional information flow and intertwining local and global features throughout the backbone.
    \item Finally, the proposed model is evaluated both on the public View-of-Delft \cite{apalffy2022} dataset and our internal dataset, surpassing previous state-of-the-art performance and demonstrating generalization across various tasks such as object detection and free-road segmentation.
\end{itemize}

\section{Related Work}
\label{sec:relatedwork}

Extensive research has been conducted on point cloud perception, including models designed for general point clouds and specifically for 4D radar point clouds. This section provides an overview of related works, with a particular focus on multi-representation models and the application of transformers in point cloud processing.

\subsection{Point Cloud Perception Models}

As discussed in \autoref{sec:introduction}, most point cloud models adopt either point-based or voxel/pillar-based representations. The inherent sparsity of point clouds, whether in raw point form or after voxelization, poses challenges for extracting informative features efficiently.

On the point-based side, PointNet \cite{Qi2017} is a pioneering work that applies a deep neural network composed of multi-layer perceptrons (MLPs) and pooling layers directly to point cloud data to extract global features. PointNet++ \cite{Li2017} extends PointNet by introducing a hierarchical structure with local PointNet modules to capture features at multiple scales,  and was used successfully for radar point cloud segmentation as well \cite{liu2022deep}. 
While these methods retain all original point-level information, high computation cost limits their receptive field, making them difficult to capture global context.

On the voxel/pillar-based side, VoxelNet \cite{Zhou2017} and PointPillars \cite{Lang2018} partition the space into a 3D/2D grid and aggregate point features within each voxel/pillar, followed by convolutions. CenterPoint \cite{Yin2020} also partitions the space into a 2D bird's-eye-view (BEV) grid and proposes a novel CenterHead, which outperforms previous anchor-based heads and makes it a widely used method for 3D object detection. SECOND \cite{Yan2018} fully exploits the sparsity of voxelized point clouds, which, inspired by \cite{Graham2017}, employs two new types of convolution: sparse convolution and submanifold sparse convolution.
These modifications to traditional convolution significantly improve training and inference efficiency. However, voxel-based methods tend to lose fine-grained spatial information during voxelization, while those information is critical for tasks requiring detailed local understanding.

\subsection{Multi-representation Models}

Multi-representation models aim to leverage the strengths of different representations, typically combining point-based and multiple voxel-based approaches to capture both local and global features. Several point-voxel dual-representation frameworks are illustrated in \autoref{fig:dual_representation_models}.

PVCNN \cite{Liu2019} introduces a voxel-based branch with convolution layers, which is fused into a parallel point-based MLP branch via devoxelization. The combined features are then further processed for segmentation and detection tasks. MVFAN \cite{Yan2023} and MUFASA \cite{Peng2024} voxelize radar point clouds into multiple 2D grids. MUFASA further includes a point-based branch to extract fine-grained features, which are fused with the voxel-based features. SMURF \cite{JLiu2023} adopts two point-based branches with kernel density estimation blocks using different bandwidths and fuses them with a pillar-based branch. 

Other works explore multi-representation from different perspectives. PVTransformer \cite{Leng2024} applies attention to point clouds before voxelization, then processes the voxelized representation with transformer layers. DRINet \cite{Ye2021} uses a cyclic architecture that iteratively switches between point and voxel representations. PVT \cite{Zhang2021} constructs voxel branches and fuses their outputs back to point branches within each PVT block.

\subsection{Transformer Applications on Point Clouds}

Transformer \cite{Vaswani2017} is a neural network architecture originally designed for sequence-to-sequence tasks and has achieved significant success in natural language processing 
and computer vision.
%
Based on attention mechanisms, transformers can model both local and global dependencies, making them well-suited for point cloud data in both point-based and voxel/pillar-based formats. In this context, each point \cite{Guo2020} or non-empty voxel/pillar \cite{Mao2021} is treated as a token. While the self-attention mechanism can be directly applied to such data, it often suffers from high computational cost due to the large number of tokens \cite{Lu2022}. To mitigate this, various methods have been proposed to improve transformer efficiency for point cloud processing.

Point Transformer \cite{Zhao2020} is the first to apply transformers to point clouds by computing self-attention over only the $k$ nearest neighbors (KNN) of each query point. However, the KNN-based approach incurs high computational cost and limits $k$ to small values, restricting the receptive field and global modeling capacity \cite{Wu2022}.
To address this, Point Transformer V3 \cite{Wu2024PTv3} introduces point cloud serialization using space-filling curves.

In voxel/pillar representations, Voxel Transformer \cite{Mao2021} applies self-attention using sparse queries and sampling modules similar to those in SECOND. RPFA-Net \cite{Xu2021} and RadarPillars \cite{Musiat2024} treat radar pillars as tokens and apply global self-attention. These methods significantly extend the receptive field compared to traditional convolutions while preserving efficiency, enabling global context modeling even in early network stages.
\section{Method}
\label{sec:method}

\begin{figure*}[t]
  \centering
  \includegraphics[width=\linewidth]{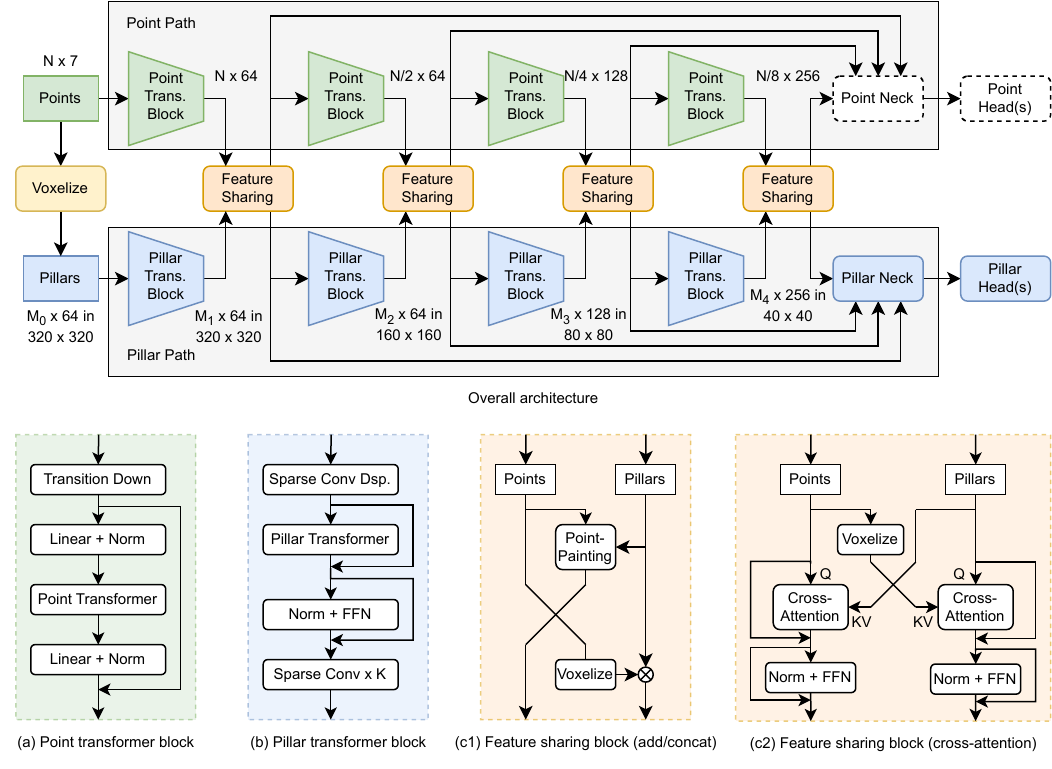}
  \caption{Model architecture. The top figure shows the overall architecture of DRIFT, which consists of a point path, a pillar path, and feature sharing blocks. The feature after block $i$: $M_i \times C_i \ \text{in} \ H_i \times W_i$ refers to a sparse pillar representation tensor with shape $M_i \times C_i$, from BEV grid size $H_i \times W_i$, where $M_i$ is the number of non-empty pillars and $C_i$ is the number of channels. (a) Point transformer block. (b) Pillar transformer block. (c1) Feature sharing block with add or concatenation fusion. (c2) Feature sharing block with cross-attention fusion.}
  \label{fig:overall_model}
\end{figure*}

The paper proposes DRIFT (Dual-Representation Inter-Fusion Transformer), a novel framework featuring end-to-end parallel point and pillar paths, with the point path aggregating fine-grained local features and the pillar path encoding coarse-grained global features. The two paths are tightly coupled with feature sharing blocks that intertwine features at multiple stages, making both paths obtain local and global information even at early stages. The overall architecture of DRIFT is shown in \autoref{fig:overall_model}. The model is implemented with transformer blocks and with efficient sparse data representations.

\subsection{Point Path}
\label{sec:point_path}

The input to the point path, which is also the overall model input, is a set of radar points \revision{accumulated from multiple ego-motion compensated scans}, with shape $N \times 7$. Here $N$ is the number of points, and the seven channels represent $x$, $y$, $z$ coordinates, radar cross-section (RCS), radial Doppler velocity relative to the ego vehicle ($v_r$), ego-motion-compensated Doppler velocity ($v_{rc}$), and a time ID ($t$) indicating the scan index following \cite{apalffy2022}.

The point path is composed of a series of point transformer blocks, designed to extract features from raw point cloud data. Each block includes a transition-down layer and point transformer encoder layers, adapted from Point Transformer \cite{Zhao2020}. The structure is illustrated in \autoref{fig:overall_model} (a). Throughout all point operations, the first three channels ($x$, $y$, $z$) are kept to maintain spatial information.

The output of each point transformer block is a point feature tensor corresponding to the downsampled points. These features are then passed to the feature sharing block, together with pillar features (see \autoref{sec:pillar_path} and \autoref{sec:feature_sharing}). The updated point features are subsequently fed into the next point transformer block. Since this work does not focus on point-based tasks, subsequent point neck and head are not implemented. 

\subsection{Pillar Path}
\label{sec:pillar_path}

The input to the pillar path is obtained via voxelization of the point cloud. The space is divided into a 2D bird's-eye-view (BEV) grid. For each pillar, the model collects the points that fall into it up to a maximum count and aggregates their features into a single vector using linear layers. Then only non-empty pillars are collected, resulting in a sparse pillar representation tensor of shape $M_0 \times C_0$, where $M_0$ is the number of non-empty pillars and $C_0$ is the feature dimension. Pillar coordinates are also stored in a tensor of shape $M_0 \times 3$, where the three channels are batch ID, and $x$, $y$ coordinates of the pillars.

The pillar path mirrors the structure of the point path, comprising several pillar transformer blocks. Each block consists of a sparse convolution downsample layer, a pillar transformer encoder, and multiple sparse convolution layers, as illustrated in \autoref{fig:overall_model} (b). Sparse convolutions \revision{\cite{Yan2018}} are used throughout the path to preserve efficiency and ensure that the sequence length \revision{(number of non-empty pillars)} remains manageable in later stages.

The pillar transformer encoder follows the standard transformer structure \cite{Vaswani2017}, as well as the design in RadarPillars \cite{Musiat2024}, and omits position embeddings, because spatial information is already encoded during voxelization. Each non-empty pillar is treated as a token, and global self-attention is applied to the features. The output is then passed through layer normalization, a feed-forward network (FFN) with GELU activation, and residual connections.

Each block outputs updated sparse pillar features and their coordinates, representing non-empty pillars in a downsampled BEV grid. These features are then passed into the feature sharing block along with the point features (see \autoref{sec:point_path} and \autoref{sec:feature_sharing}). All pillar outputs from the feature sharing blocks are then processed through a feature pyramid network \cite{Lin2017}, followed by a pillar head, as shown in \autoref{fig:overall_model}.

\subsection{Feature Sharing Blocks}
\label{sec:feature_sharing}

Feature sharing blocks enable bi-directional information exchange between the point and pillar paths, allowing the model to leverage both fine-grained and global representations in both paths. These blocks are inserted at multiple stages and represent a key contribution of this work. Three fusion strategies are implemented: addition, concatenation, and cross-attention. Addition and concatenation methods offer simple and efficient inter-fusion, while the cross-attention method is more complex but captures more intricate relationships between the point and pillar features. The structures of the various feature sharing blocks are illustrated in \autoref{fig:overall_model} (c1) (for add or concatenation fusion) and \autoref{fig:overall_model} (c2) (for cross-attention fusion), respectively.

\subsubsection{Add \& Concatenation Fusion}
\label{sec:add_concat_fusion}

To transfer pillar features to the point path, the idea of PointPainting \cite{vora2020pointpainting} is adopted, but instead of image features, the painting of the points is now performed with pillars. First, it computes the 2D BEV grid coordinates of each point, retrieves the corresponding pillar features, and adds or concatenates the pillar and point features. A linear layer is then applied to adjust the channel dimension. 

To transfer the point features to the pillar path, the point features are voxelized and then added or concatenated to the existing pillar features. A linear layer adjusts the feature dimension afterward. For pillars without corresponding point features, a learnable token is introduced as a placeholder.

\subsubsection{Cross-Attention Fusion}
\label{sec:cross_attention_fusion}

As an alternative to the classical add and concatenation methods, a cross-attention mechanism is also proposed to share features between the point and voxel path. In pillar-to-point direction, point features serve as queries, and non-empty pillar features serve as keys and values. For the reverse direction, it first voxelizes the point features and uses the resulting features as keys and values, with non-empty pillar features as queries. Layer normalization and FFN layers with residual connections are then applied after attention.

\begin{table*}[t]
\centering
\caption{Comparison of VoD validation set results. DRIFT uses cross-attention feature sharing. Pre-training is performed by perciv-scenes-2 dataset. The best results of each configuration (with and without pre-training) are bold. Results marked with \dag{} are inherited from \cite{JLiu2023}, results marked with * comes from re-implemented methods.}
\label{tab:overall_vod}
\begin{threeparttable}
\begin{tabular}{l|cccc|cccc}
\toprule
\multirow{2}{*}{Method}        & \multicolumn{4}{c|}{Entire Area} & \multicolumn{4}{c}{Driving Corridor} \\
                                & mAP $\uparrow$ & Car $\uparrow$ & Ped. $\uparrow$ & Cyc. $\uparrow$ & mAP $\uparrow$ & Car $\uparrow$ & Ped. $\uparrow$ & Cyc. $\uparrow$ \\
\midrule
PointPillars\tnote{\dag} \hspace{0.1em} \cite{Lang2018}    & 45.2  & 37.1  & 35.0   & 63.4   &  67.5   &  70.2  &  47.2   &  85.1   \\
CenterPoint\tnote{\dag} \hspace{0.1em} \cite{Yin2020}      & 45.4  & 32.7  & 38.0   & 65.5   &  65.1   &  62.0  &  48.2   &  85.5   \\
CenterPoint\tnote{*} \hspace{0.1em} \cite{Yin2020}         & 47.4  & 39.9  & 37.8  & 64.7  & 67.4  & 71.3  & 47.3  & 83.7  \\
PillarNeXt \cite{Li2023}        & 42.2  & 30.8  & 33.1   & 62.8   &  63.6   &  66.7  &  39.0   &  85.1   \\
RadarPillars \cite{Musiat2024}  & 50.7  & 41.1  & 38.6   & 72.6   &  70.5   &  71.1  & 52.3   &  87.9   \\
MVFAN (single scan) \cite{Yan2023}            & 39.4  & 34.1  & 27.3   & 57.1   &  64.4   &  69.8  &  38.7   &  84.9   \\
MUFASA \cite{Peng2024}          & 50.2  & \textbf{43.1}  & 39.0   & 68.7   &  69.4   &  71.9  &  47.4   &  89.0   \\
SMURF \cite{JLiu2023}           & 51.0  & 42.3  & 39.1   & 71.5   &  69.7   &  71.7  &  50.5   &  86.9   \\
DRIFT (attention) (proposed)        & \textbf{52.6}  & 41.4  & \textbf{42.2}   & \textbf{74.3}   &  \textbf{71.5}   &  \textbf{72.4}  &  \textbf{52.9}   &  \textbf{89.3}   \\
\midrule
CenterPoint (pre-trained)\tnote{*} \hspace{0.1em} \cite{Yin2020}      & 47.7  & 39.7  & 35.5  & 67.9  & 66.4  & 70.3  & 45.7  & 83.1  \\
DRIFT (attention, pre-trained) (proposed)        & \textbf{53.1}  & \textbf{41.2}  & \textbf{43.6}  & \textbf{74.5}  & \textbf{72.4}  & \textbf{77.0}  & \textbf{52.5}  & \textbf{87.8}  \\
\bottomrule
\end{tabular}
\end{threeparttable}
\end{table*}

\section{Experiments}
\label{sec:experiments}

The method is evaluated on two tasks: object detection and free-road segmentation. The object detection task predicts 3D bounding boxes for vehicles, pedestrians, and cyclists, whereas the free‑road task identifies unoccupied drivable areas. 

\subsection{Datasets and Evaluation Metrics}
\label{sec:datasets}

Two datasets are used for evaluation: the View-of-Delft (VoD) dataset \cite{apalffy2022} for object detection and the internal perciv-scenes-2 dataset for both object detection and free-road segmentation. The VoD dataset contains 8.7k frames of synchronized and calibrated LiDAR, camera, and 4D radar data with 123k annotated 3D object bounding boxes, recorded in Delft, the Netherlands. The perciv-scenes-2 dataset is an internal dataset from us, comprising 130k frames of data with 2.7M object bounding boxes, recorded on German highways and in the Rotterdam-The Hague Metropolitan Region (including Delft), using the same radar sensor as VoD. Unlike VoD, perciv-scenes-2 has not only object bounding box annotations, but also BEV mask annotations to mark free space in the scene.

For the VoD dataset, five-scan accumulated radar point clouds are employed. The object detection result is evaluated with the official VoD script, reporting per‑class average precision (AP) and mean AP (mAP) over the entire scene ($\text{AP}_\text{all}$) and the driving corridor (a narrow and short-range rectangular area that is the safety critical region, see \cite{apalffy2022}) ($\text{AP}_\text{roi}$). \revision{In this dataset, DRIFT is compared with state-of-the-art methods, and ablation studies are conducted to analyze the contribution of individual components.}
Intersection over union (IoU) thresholds are 0.5 for cars and 0.25 for pedestrians and cyclists.

For the internal perciv-scenes-2 dataset, the object detection task takes the same input format and classes as VoD. The evaluation follows the nuScenes \cite{nuscenes2019} metrics: mAP and the nuScenes detection score (NDS), which is a weighted average score of mAP and the mean average error of translation, scale, orientation, velocity, and box attribute of true positives. Since perciv-scenes-2 lacks attribute labels, the mean average attribute error is fixed to 1. In the free-road segmentation task, BEV occupancy is first predicted, after which a ray‑casting algorithm identifies free‑road cells by tracing from the sensor to the nearest occupied cell in predefined directions. Performance is measured with IoU for free-road area ($\text{IoU}_\text{free}$) and binarized occupancy map ($\text{IoU}_\text{occ}$). \revision{For this internal dataset, we use CenterPoint as baseline, as it is the best-performing method with an open-source implementation.}

\subsection{Implementation Details}
\label{sec:implementation}

Implementation of the model is based on PyTorch within the MMDetection3D framework \cite{mmdet3d2020}.

Point clouds are cropped to 0 to 51.2 meters on the x-axis, -25.6 to 25.6 meters on the y-axis, and -3 to 2 meters on the z-axis with a voxel size of (0.16 m, 0.16 m), producing a (320, 320) grid. For the free-road segmentation task, voxel size is (0.2 m, 0.2 m), so grid size is (256, 256).

The point and pillar paths both consist of 4 transformer blocks, with output feature dimensions of [64, 64, 128, 256]. The transition-down strides across the four stages are [1, 2, 2, 2], and the number of sparse convolution layers in each pillar transformer block are [1, 3, 5, 5].

The outputs from the last three pillar stages are fed into a Feature Pyramid Network neck. For object detection, a CenterPoint head \cite{Yin2020} is used. For free-road segmentation, an 8-layer convolutional head with a transposed convolutional layer after the $4^{\text{th}}$ layer is adopted.

During training, augmentations are applied to the detection task, including random flipping, rotation, scaling, and ground truth database sampling \cite{Yan2018}. The model is optimized using AdamW with an initial learning rate of $1\times10^{-4}$ and a weight decay of 0.01. The model is trained for 80 epochs on VoD and 16 epochs on perciv-scenes-2, using an NVIDIA RTX 4090 GPU. 
Validation is performed every four epochs on VoD and every epoch on perciv-scenes-2. The epoch with the best validation $\text{mAP}_\text{all}$ is reported for each run.


\begin{table*}[t]
\centering
\caption{Comparison of object detection and free-road segmentation results on perciv-scenes-2 validation set.}
\label{tab:overall_perciv}
\begin{tabular}{l|ccccc|cc}
\toprule
\multirow{2}{*}{Method}        & \multicolumn{5}{c|}{Object Detection} & \multicolumn{2}{c}{Free-road} \\
                                & mAP $\uparrow$ & NDS $\uparrow$ & 
                                $\text{AP}_\text{car}$$\uparrow$ & $\text{AP}_\text{ped.}$$\uparrow$& $\text{AP}_\text{cyc.}$$\uparrow$ &
                                $\text{IoU}_\text{free}$$\uparrow$ & $\text{IoU}_\text{occ}$$\uparrow$ \\
\midrule
CenterPoint \cite{Yin2020}      & 51.8  & 50.9  & 55.5  & 45.2  & 54.6  & 71.5  & 28.6   \\
DRIFT (attention)               & \textbf{55.2}  & \textbf{52.7}  & \textbf{56.0}  & \textbf{51.8}  & \textbf{57.8}  & \textbf{73.3}  & \textbf{29.7}  \\
\bottomrule
\end{tabular}
\end{table*}

\subsection{Main Results}
\label{sec:main_results}

A comprehensive comparison of the proposed method (with cross-attention feature sharing) with existing point cloud object detection techniques is performed on the VoD validation set. This covers models originally designed for LiDAR data \cite{Lang2018,Yin2020}, radar‑specific models \cite{Musiat2024}, as well as multi‑representation approaches \cite{Yan2023,Peng2024,JLiu2023}. The results are shown in \autoref{tab:overall_vod}.

Results show that DRIFT achieves state-of-the-art performance in the VoD dataset, with an mAP of 52.6\% in the entire area and 71.5\% in the driving corridor. The proposed method outperforms all other methods in terms of mAP, especially for the pedestrian and cyclist detection results, where it achieves 42.2\% / 74.3\% on the entire area and 52.9\% / 89.3\% on the driving corridor. The high performance on these two classes indicates the benefit of the dual path design, as detecting these smaller classes often requires the combination of fine-grained local information and global context.

\revision{Transformer-based models often require substantially more training data, and the amount of data in VoD is limited. Therefore, we also pre-trained the model on our larger perciv-scenes-2 dataset.} This further improves performance to 53.1\% $\text{mAP}_\text{all}$ and 72.4\% $\text{mAP}_\text{roi}$, whereas the CenterPoint baseline shows no improvement. This strong dependence on training data suggests that the proposed method may benefit from even larger-scale datasets.

\autoref{tab:overall_perciv} summarizes results from both object detection and free-road segmentation tasks on the perciv-scenes-2 validation set. When doing free-road segmentation, the CenterPoint head is replaced with the same occupancy head used in the proposed method. The object detection task requires both local and global context, and the free-road segmentation task requires more local context. The proposed method surpasses CenterPoint on both tasks, confirming its versatility across tasks requiring different balances of local and global context.

\begin{figure*}[t]
    \centering
    \begin{subfigure}{0.32\linewidth}
        \centering
        \includegraphics[width=\textwidth, trim=100 100 220 200, clip]{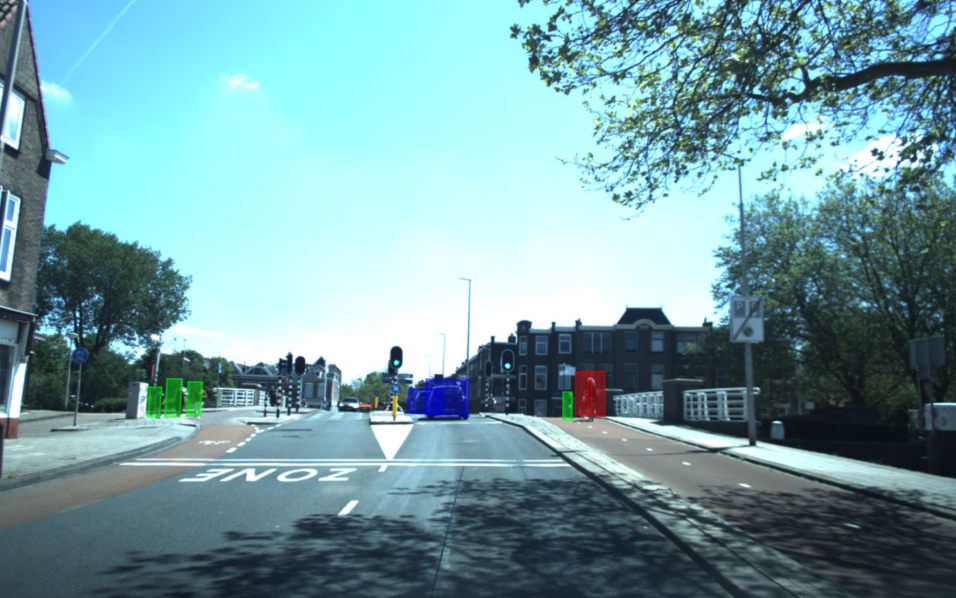}
        \caption{Ground truth 1}
        \label{fig:gt_00030}
    \end{subfigure}
    \begin{subfigure}{0.32\linewidth}
        \centering
        \includegraphics[width=\textwidth, trim=100 100 220 200, clip]{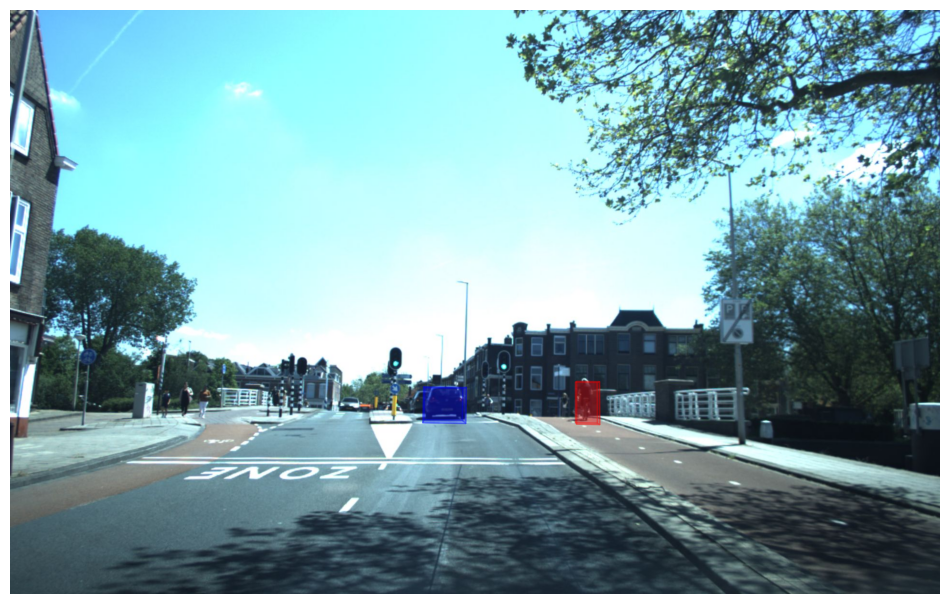}
        \caption{CenterPoint prediction 1}
        \label{fig:centerpoint_00030}
    \end{subfigure}
    \begin{subfigure}{0.32\linewidth}
        \centering
        \includegraphics[width=\textwidth, trim=100 100 220 200, clip]{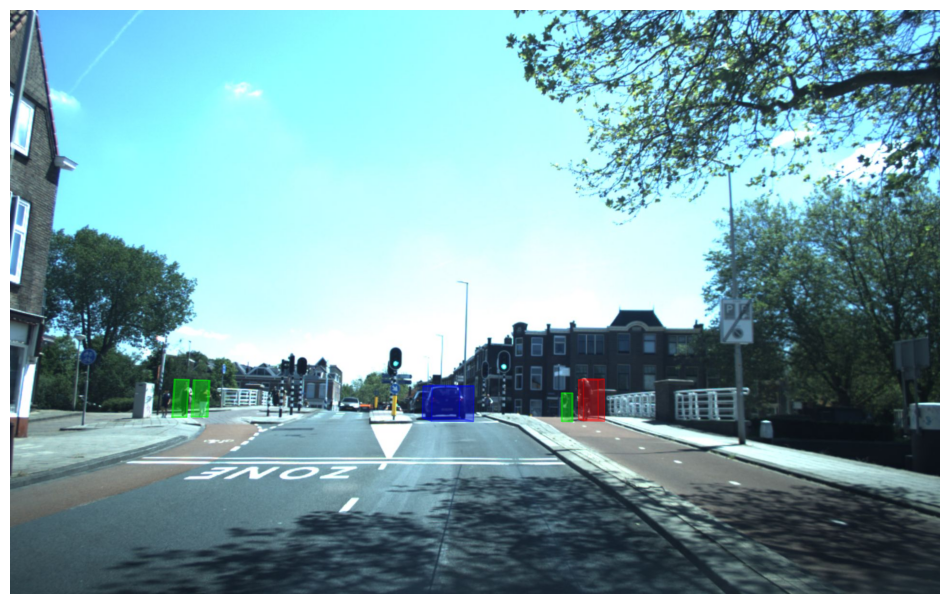}
        \caption{Proposed model's prediction 1}
        \label{fig:ours_00030}
    \end{subfigure}

    \vspace{0.01\textwidth}
    
    \begin{subfigure}{0.32\linewidth}
        \centering
        \includegraphics[width=\textwidth, trim=0 0 0 150, clip]{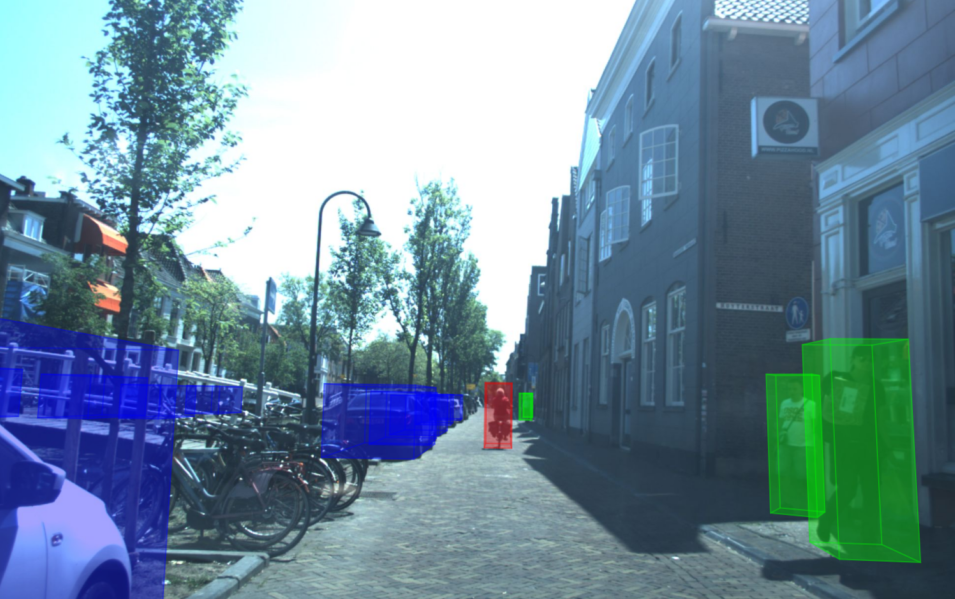}
        \caption{Ground truth 2}
        \label{fig:gt_00302}
    \end{subfigure}
    \begin{subfigure}{0.32\linewidth}
        \centering
        \includegraphics[width=\textwidth, trim=0 0 0 150, clip]{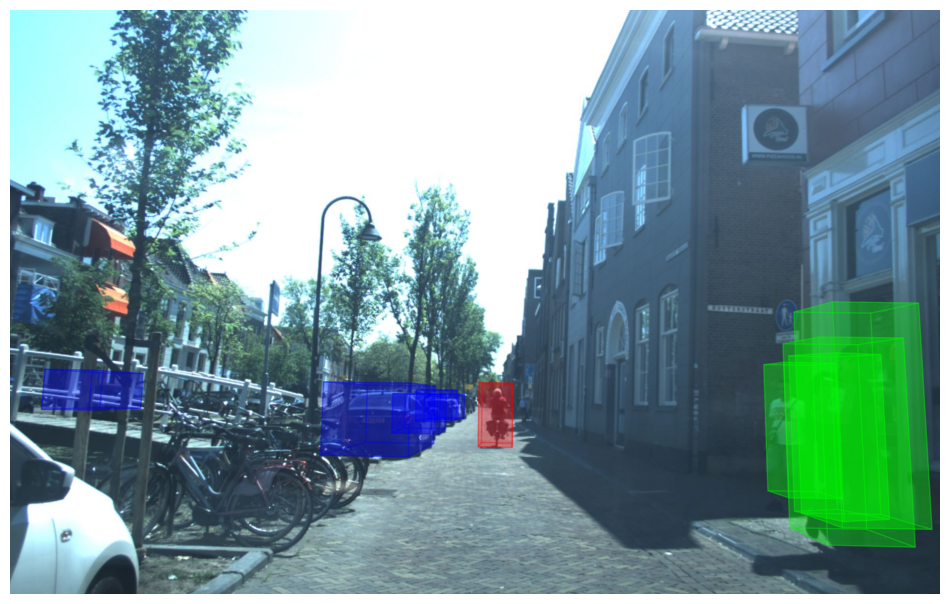}
        \caption{CenterPoint prediction 2}
        \label{fig:centerpoint_00302}
    \end{subfigure}
    \begin{subfigure}{0.32\linewidth}
        \centering
        \includegraphics[width=\textwidth, trim=0 0 0 150, clip]{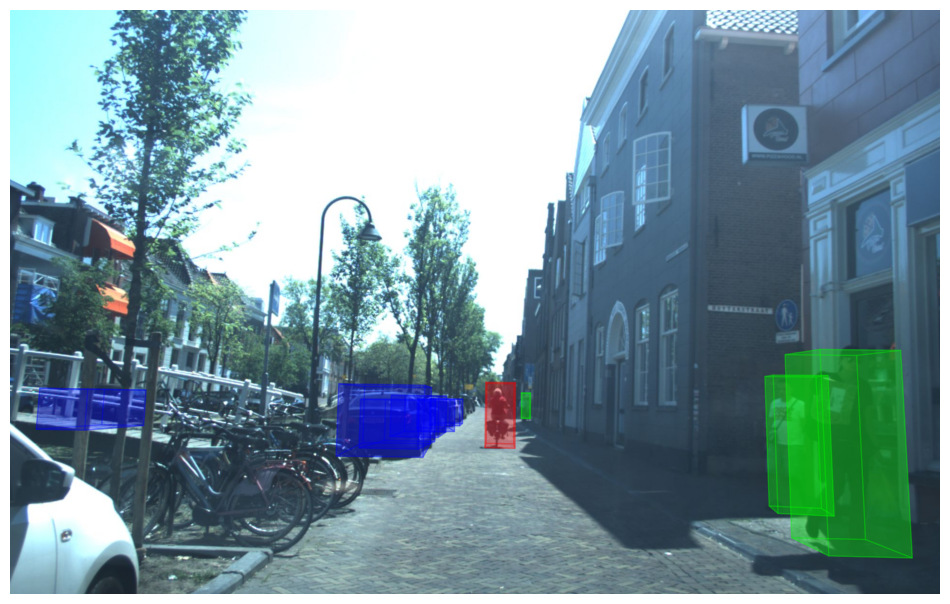}
        \caption{Proposed model's prediction 2}
        \label{fig:ours_00302}
    \end{subfigure}

    \caption{Visualization results of ground truth, CenterPoint, and DRIFT (proposed) on the VoD validation set. Blue/green/red boxes indicate car, pedestrian, and cyclist detections, respectively. All images are trimmed.}
    \label{fig:vis_vod}
\end{figure*}

\begin{figure}[t]
    \centering
    \begin{subfigure}{0.45\linewidth}
        \centering
        \includegraphics[width=\textwidth, trim=0 0 0 0, clip]{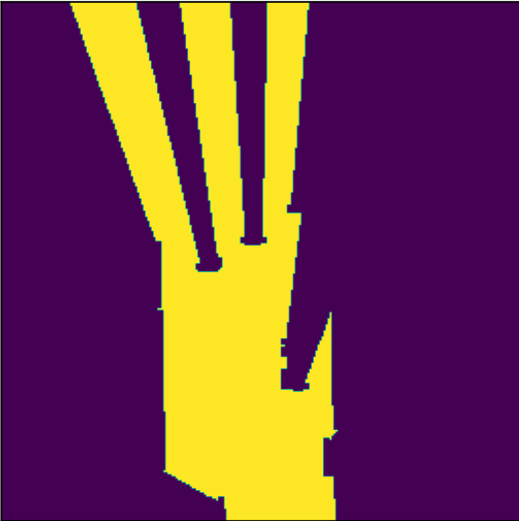}
        \caption{Ground truth 1}
        \label{fig:occ_gt_1}
    \end{subfigure}
    \begin{subfigure}{0.45\linewidth}
        \centering
        \includegraphics[width=\textwidth, trim=0 0 0 0, clip]{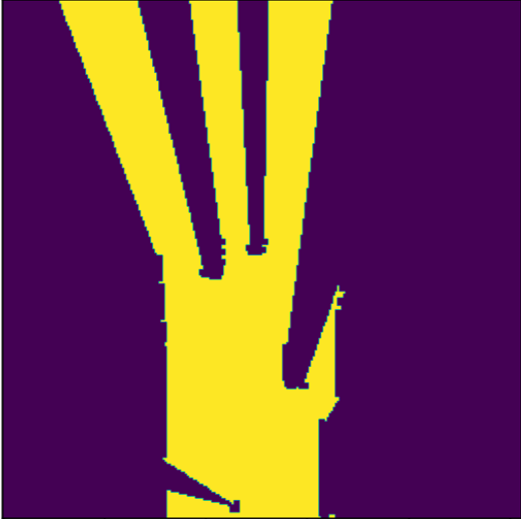}
        \caption{Prediction 1}
        \label{fig:occ_pred_1}
    \end{subfigure}

    \vspace{0.01\textwidth}
    
    \begin{subfigure}{0.45\linewidth}
        \centering
        \includegraphics[width=\textwidth, trim=0 0 0 0, clip]{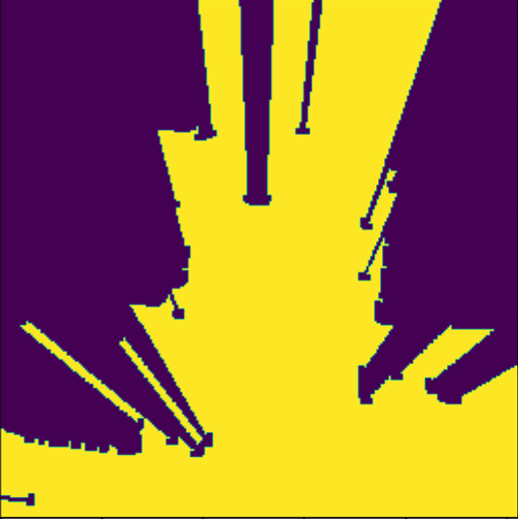}
        \caption{Ground truth 2}
        \label{fig:occ_gt_2}
    \end{subfigure}
    \begin{subfigure}{0.45\linewidth}
        \centering
        \includegraphics[width=\textwidth, trim=0 0 0 0, clip]{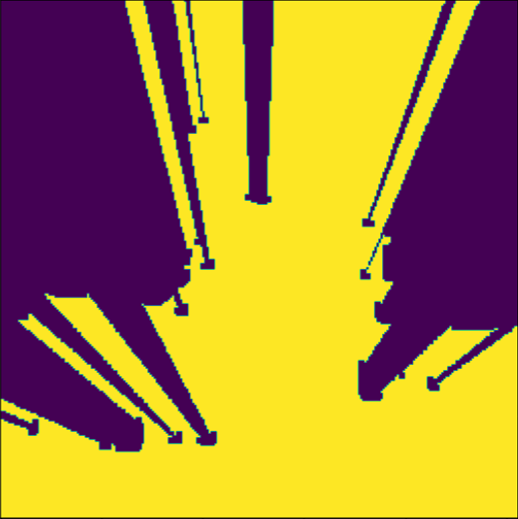}
        \caption{Prediction 2}
        \label{fig:occ_pred_2}
    \end{subfigure}

    \caption{Visualization of free-road ground truth and prediction on perciv-scenes-2 validation set on BEV. Yellow segment indicate free-road region.}
    \label{fig:vis_occ}
\end{figure}

\autoref{fig:vis_vod} visualizes the performance of the proposed model compared with the CenterPoint baseline on the VoD validation set. The first row shows results in a typical Dutch urban road scene. Compared to the baseline, the proposed method achieves a longer detection range and performs especially well on smaller objects such as pedestrians. The second row shows a narrow and complex road scene. CenterPoint mixes up noise points and pedestrian points at a near location, resulting in false positive pedestrian detections. It also fails to detect a distant pedestrian that only reflects a few radar points. In contrast, the proposed method successfully detects correct objects in both cases. \autoref{fig:vis_occ} visualizes the performance of the proposed model for free-road segmentation on the perciv-scenes-2 validation set, indicating the effectiveness of the model on this task.

\revision{During validation, DRIFT requires 4.92 GB / 7.74 GB of GPU memory, has 6.63 M / 7.45 M parameters, and achieves 16.4 ms / 20.0 ms inference latency with addition / cross-attention feature sharing, respectively, demonstrating real-time inference capability.}

\subsection{Ablation Study}
\label{sec:ablation_study}

The effectiveness of various components in the proposed model is evaluated through ablation studies on the VoD validation set. These include the dual-path representation structure, transformer blocks, and feature sharing blocks. To reduce randomness and assess statistical significance, each configuration is trained three times using different random seeds, and the mean and sample standard deviation are reported.

\begin{table}[htbp]
\centering
\caption{Ablation study on the two data representation paths and transformer blocks.}
\label{tab:dual_path_ablation}
\begin{tabular}{l|cc}
\toprule
Configuration                      & $\text{mAP}_\text{all}$ $\uparrow$  & $\text{mAP}_\text{roi}$ $\uparrow$  \\
\midrule
Pillar Path (no trans.)             & 46.8$\pm$0.6  & 67.5$\pm$0.1 \\
Pillar Path (no trans., double channel)      & 47.5$\pm$0.6  & 68.1$\pm$2.2 \\
Pillar Path (trans. in 1st block)   & 48.6$\pm$1.3  & 69.3$\pm$1.0 \\
Pillar Path (trans. in all blocks)  & 49.6$\pm$0.7  & 69.3$\pm$0.8 \\
Pillar Path (trans. in all blocks, double channel) & 50.0$\pm$0.5  & 69.1$\pm$0.8 \\
Dual Path (trans. in 1st blocks)    & 49.2$\pm$0.8  & 68.9$\pm$0.5 \\
Dual Path (trans. in all blocks)    & \textbf{51.5$\pm$1.1}  & \textbf{70.6$\pm$0.8} \\
\bottomrule
\end{tabular}
\end{table}

To evaluate the dual-path transformer structure, experiments are conducted by retaining only the pillar path and/or removing the transformer layers, and results are shown in \autoref{tab:dual_path_ablation}. For some configurations, the channel size is doubled, making the inference time and memory similar to DRIFT, and the number of parameters even much larger than DRIFT. Results indicate that both the dual-path design and the use of transformer mechanisms in both paths and all blocks contribute positively to performance, \revision{rather than the gains coming from simply scaling up model size.}

Further experiments are conducted to assess different types of feature sharing methods on the VoD validation set. As shown in the first three rows of \autoref{tab:feature_sharing}, when using the same type of feature sharing method for both point-to-pillar and pillar-to-point directions, the cross-attention method yields the best results. The final two rows analyze the contribution of the cross-attention method in each direction. Pillar-to-point cross-attention has a greater effect than point-to-pillar one on performance, \revision{which is likely due to the information loss introduced during voxelization in point-to-pillar fusion, a trade-off made for computational efficiency,} though bi-directional cross-attention remains essential for achieving the best results.

\begin{table}[htbp]
\centering
\caption{Comparison of different types of feature sharing methods.}
\label{tab:feature_sharing}
\begin{tabular}{cc|cc}
\toprule
\multicolumn{2}{c|}{Fusion Configuration} & \multirow{2}{*}{$\text{mAP}_\text{all}$ $\uparrow$}  & \multirow{2}{*}{$\text{mAP}_\text{roi}$ $\uparrow$}  \\
Point to Pillar    & Pillar to Point  \\
\midrule
Add             & Add           & 49.9$\pm$0.7  & 69.7$\pm$0.5 \\
Concat          & Concat        & 49.4$\pm$0.9  & 69.8$\pm$0.8 \\
Attention       & Attention     & \textbf{51.5$\pm$1.1}  & \textbf{70.6$\pm$0.8} \\
\midrule
Add             & Attention     & 50.7$\pm$0.2  & 69.6$\pm$1.1 \\
Attention       & Add           & 50.2$\pm$1.2  & 69.6$\pm$0.6 \\
\bottomrule
\end{tabular}
\end{table}


\begin{table}[htbp]
\centering
\caption{Ablation study on feature sharing blocks.}
\label{tab:feature_sharing_ablation}
\begin{tabular}{l|cc}
\toprule
Fusion Configuration   & $\text{mAP}_\text{all}$ $\uparrow$  & $\text{mAP}_\text{roi}$ $\uparrow$ \\
\midrule
None (pillar path only) & 49.0$\pm$0.4  & 69.4$\pm$0.6 \\
Point to Pillar Only    & 49.4$\pm$0.2  & 68.8$\pm$0.7 \\
First Block Only        & 49.4$\pm$0.4  & 69.9$\pm$0.4 \\
Last Block Only         & 49.9$\pm$0.2  & 69.1$\pm$0.2 \\
All Blocks              & \textbf{51.5$\pm$1.1}  & \textbf{70.6$\pm$0.8} \\
\bottomrule
\end{tabular}
\end{table}

To validate the importance of multi-stage and bi-directional feature sharing, an additional ablation study is conducted with different feature sharing configurations. The results are shown in \autoref{tab:feature_sharing_ablation}, \revision{indicating that the point path can effectively utilize global information from the pillar path when extracting point-wise local features, while multi-stage feature sharing ensures sufficient information exchange between the two paths,} highlighting the benefit of applying feature sharing at all stages and in both directions.


\section{Conclusion}
\label{sec:conclusion}

This paper presents DRIFT, a novel 4D radar point cloud perception framework that integrates end-to-end parallel point and pillar representation paths, inter-fused with bi-directional feature sharing blocks across multiple stages. Attention mechanisms are incorporated in both the point and pillar paths, as well as in the feature sharing layers, to strengthen the point path’s capacity for capturing fine-grained local details and the pillar path’s capability for modeling global context. The feature sharing blocks, which are the main contribution of this paper, serve as a key component, enabling effective information exchange between the two paths, allowing the point-based representation to benefit from the contextual abstraction of the pillar path, and vice versa.

Extensive experiments on the View-of-Delft dataset and the internal-scene dataset demonstrate the effectiveness of the proposed architecture. DRIFT surpasses the previous state-of-the-art performance across multiple benchmarks and tasks, particularly excelling in the detection of small and distant objects such as pedestrians and cyclists, where scenarios are especially challenging for radar-based perception due to sparse and noisy inputs. Ablation studies further verify the contributions of each design component, including the dual-path architecture, the use of attention in both branches and feature sharing blocks, as well as the multi-stage bi-directional feature sharing strategy.

Further research could explore the role of each component in the architecture, such as the same blocks in different stages, and the impact of different attention mechanisms. Evaluating the model on other tasks, such as point-wise semantic segmentation or instance segmentation, may also help further validate the flexibility and effectiveness of the proposed design. Finally, evaluating DRIFT in broader domains, such as full-stack autonomous driving systems, robot navigation, or aerial perception, may offer deeper insights into the generalization capability of the framework.

\bibliographystyle{IEEEtran}
\bibliography{main}

\end{document}